\documentclass[a4paper, 10pt, conference]{ieeeconf}      

\IEEEoverridecommandlockouts                              

\usepackage{amssymb}
\usepackage{amsmath}
\usepackage{bm}
\usepackage{balance}
\usepackage{lineno}
\usepackage{tabularx}
\usepackage{booktabs}

\usepackage{graphicx}      
\graphicspath{{figures/}}
\usepackage{algorithm}
\usepackage{algpseudocode}
\usepackage{hyperref}

\usepackage[T1]{fontenc}
\usepackage{xcolor}

\usepackage[capitalise]{cleveref}

\title{\LARGE \bf
Meta-Reinforcement Learning for Adaptive Control of Second Order Systems
}

\author{Daniel G. McClement$^{1}$, Nathan P. Lawrence,$^{2}$, Michael G. Forbes$^{3}$,\\ Philip D. Loewen$^{2}$, Johan U. Backstr{\"o}m$^{4}$, R. Bhushan Gopaluni$^{1}$
\thanks{$^{1}$Department of Chemical and Biological Engineering, University of British Columbia, Vancouver, BC Canada}
\thanks{$^{2}$Department of Mathematics, University of British Columbia, Vancouver BC, Canada}
\thanks{$^{3}$Honeywell Process Solutions, North Vancouver, BC Canada}
\thanks{$^{4}$Backstrom Systems Engineering Ltd., North Vancouver, BC Canada}
\thanks{\copyright 2022 the authors. This work has been accepted to AdCONIP 2022. We gratefully acknowledge the financial support from Natural Sciences and Engineering Research Council of Canada (NSERC) and Honeywell Connected Plant.}
}

\begin{document}

\maketitle
\thispagestyle{empty}
\pagestyle{empty}





\begin{abstract}                


Meta-learning is a branch of machine learning which aims to synthesize data from a distribution of related tasks to efficiently solve new ones. In process control, many systems have similar and well-understood dynamics, which suggests it is feasible to create a generalizable controller through meta-learning. In this work, we formulate a meta reinforcement learning (meta-RL) control strategy that takes advantage of known, offline information for training, such as a model structure. The meta-RL agent is trained over a distribution of model parameters, rather than a single model, enabling the agent to automatically adapt to changes in the process dynamics while maintaining performance. A key design element is the ability to leverage model-based information offline during training, while maintaining a model-free policy structure for interacting with new environments. Our previous work has demonstrated how this approach can be applied to the industrially-relevant problem of tuning proportional-integral controllers to control first order processes. In this work, we briefly reintroduce our methodology and demonstrate how it can be extended to proportional-integral-derivative controllers and second order systems.
\end{abstract}



\section{Introduction}

\emph{Meta-learning}, or ``learning to learn'', is an active area of research in which the objective is to learn an underlying structure governing a distribution of possible tasks \cite{huisman2021survey}. This framework can be applied to reinforcement learning (RL) for learning a generalized goal-oriented ``policy'' over a large number of stochastic environments \cite{finn2017model}. In process control applications, meta reinforcement learning (meta-RL) is appealing because many systems have similar dynamics or a known structure \cite{optimalpid}, which suggests training a policy over a distribution of systems could enable more rapid online deployment without specifying a system-specific model. Moreover, extensive online learning is impractical for training over a large number of systems; by focusing on learning a underlying structure across systems with different process dynamics, we can more readily adapt to a new system.\par 

This paper proposes a method for improving the \emph{online} sample efficiency of RL agents. Our approach is to train a meta-RL agent \emph{offline} by exposing it to a broad distribution of different dynamics (or ``tasks''). The agent synthesizes its experience from different environments to quickly learn an optimal policy for its present environment. The basis for our strategy is a recurrent neural network (RNN) policy. The hidden state of the RNN serves as an encoding of the system dynamics, which provides the network with ``context'' for its policy. We aim to use this framework to develop a universal controller which can quickly adapt to effectively control any process rather than a single task. The training is performed completely offline and the result is a single RL agent that can quickly adapt its policy to a new environment in a model-free fashion. \par

We apply this general method to the industrially-relevant problem of autonomous controller tuning and adaptation. We show how our trained agent can adaptively fine-tune proportional-integral-derivative (PID) controller parameters when the underlying dynamics drift or are not precisely known. Perhaps the most appealing consequence of this method is that it removes the need to accommodate a training algorithm on a system-by-system basis --- for example, through extensive online training or transfer learning, hyperparameter tuning, or system identification --- because the adaptive policy is pre-computed and represented in a single model. \par

This paper extends our previous work which focused only on systems which could be modeled as first order plus time delay and PI controllers \cite{mcclement2022meta2}. In particular, in \cref{sec:algorithm} we extend our framework to second order systems and PID controllers. \cref{sec:results} provides new simulation studies and analysis of the trained meta-RL agent.



\section{Background}
\label{sec:background}

\subsection{Reinforcement learning}
\label{subsec:RL}

In this section, we give a brief overview of deep RL. We refer the reader to \cite{nian2020review, spielberg2019toward}, for tutorial overviews of deep RL with applications to process control. We use the standard RL terminology that can be found in \cite{sutton2018reinforcement}. \cite{huisman2021survey} gives a unified survey of deep meta-learning.\par 
The RL framework consists of an \emph{agent} and an \emph{environment}. For each \emph{state} $s_t \in \mathcal{S}$ (the state-space) the agent encounters, it takes some \emph{action} $a_t \in \mathcal{A}$ (the action-space), leading to a new state $s_{t+1}$. The action is chosen according to a conditional probability distribution $\pi$ called a \emph{policy}; we denote this relationship by $a_t \sim \pi(a_t | s_t)$. Although the system dynamics are not necessarily known, we assume they can be described as a Markov decision process (MDP) with initial distribution $p(s_0)$ and transition probability $p(s_{t+1} | s_t, a_t)$. A state-space model in control is a special case of an MDP, where the states are the (minimal realization) vector that characterizes the system, while the actions are the control inputs. At each time step, a bounded scalar \emph{cost} $c_t = c(s_t, a_t)$ is evaluated. The cost function describes the desirability of a state-action pair and defining it is a key part of the design process. The overall objective, however, is the expected long-term cost. In terms of a user-specified discount factor $0<\gamma<1$, the optimization problem of interest becomes
\begin{equation}
\begin{aligned}
    &\text{minimize} && J(\theta) = \mathbb{E}_{h \sim p^{\pi_{\theta}}(\cdot)}\left[ \sum_{t=1}^{\infty} \gamma^{t-1}c(s_t,\pi_{\theta}(s_t)) \middle| s_0 \right]\\
    &\text{over all} && \theta \in \mathbb{R}^{n}.
\end{aligned}
\label{eq:RLobjective}
\end{equation}
In this problem, $h \sim p^\pi$ refers to a typical trajectory $h~=~(s_0, a_0, c_0, , \ldots, s_N, a_N, c_N)$ generated by the policy $\pi$ with subsequent states distributed according to $p$. Within the space of all possible policies, we optimize over a parameterized subset whose members are denoted $\pi_{\theta}$. We use $\theta$ as a generic vector of parameters.\par 
Common approaches to solving Problem~(\ref{eq:RLobjective}) involve techniques based on $Q$-learning (value-based methods) and the policy gradient theorem (policy-based methods) \cite{sutton2018reinforcement}, or a combination of both called \emph{actor-critic} methods \cite{konda2000actor}. Closely-related functions to $J$ are the $Q$-function (state-action value function) and value function, respectively:
\begin{align}
    Q(s_t, a_t) &= \mathbb{E}_{h \sim p^{\pi}(\cdot)}\left[ \sum_{k = t}^{\infty} \gamma^{t-1}c(s_k,a_k) \middle| s_t, a_t \right]\label{eq:Qfunc}\\
    V(s_t) &= \mathbb{E}_{h \sim p^{\pi}(\cdot)}\left[ \sum_{k = t}^{\infty} \gamma^{t-1}c(s_k,a_k) \middle| s_t \right].\label{eq:Valuefunc}
\end{align}
The \emph{advantage function} is then $A(s, a) = Q(s, a) - V(s)$. These functions help form the basis for deep RL algorithms, that is, algorithms that use deep neural networks to solve RL tasks. Deep neural networks are a flexible form of function approximators, well-suited for learning complex control laws. Moreover, function approximation methods make RL problems tractable in continuous state and action spaces \cite{silver2014deterministic, sutton2000policy}.\par 

A standard approach to solving Problem~(\ref{eq:RLobjective}) uses gradient descent:
\begin{equation}
\theta
\leftarrow \theta - \alpha\nabla J(\theta),
\label{eq:PolicyGradient_Iteration}
\end{equation}
where $\alpha > 0$ is a step-size parameter.\par 
Analytic expressions for such a gradient exist for both stochastic and deterministic policies \cite{sutton2018reinforcement, silver2014deterministic}. However, in practice, approximations are necessary. We use the proximal policy optimization (PPO) reinforcement learning algorithm to optimize Problem~(\ref{eq:RLobjective}) using a surrogate objective to compute \eqref{eq:PolicyGradient_Iteration} based on the advantage function \cite{schulman2017proximal}. We use PPO because of its sample efficiency and robustness to the choice of model hyperparameters. Full implementation and technical details can be found in \cite{mcclement2022meta2, schulman2017proximal}. \par

\subsection{Meta reinforcement learning}

\begin{figure}
\begin{minipage}[]{\linewidth}
\begin{center}
    \includegraphics[width=\textwidth]{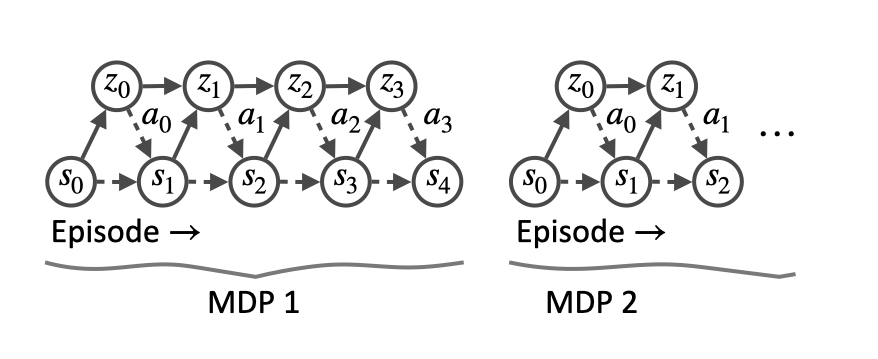}
    \caption{A diagram of the meta-RL agent's interactions with the task distribution. Each MDP represents a different task sampled from $p(\mathcal{T})$.}
    \label{fig:RL2}
\end{center}
\end{minipage}
\end{figure}

While the RL framework mentioned above can achieve impressive results in a wide range of domains, it is formulated for a single MDP. In contrast, meta-RL aims to generalize agents to a distribution of MDPs. Formally, a single MDP can be characterized by a tuple $\mathcal{T} = (\mathcal{S}, \mathcal{A}, p, c, \gamma)$; in contrast, meta-RL tackles an optimization problem over a distribution $p_{\text{meta}}(\mathcal{T})$ of MDPs. Therefore, in the meta-RL terminology, a ``task'' is simply all the components comprising a single RL problem. The problem of interest in the meta-RL setting is a generalization of the standard RL objective in Problem~(\ref{eq:RLobjective}) \cite{huisman2021survey}:
\begin{equation}
\begin{aligned}
    &\text{minimize} && J_{\text{meta}}(\bm{\Theta}) = \mathbb{E}_{\mathcal{T} \sim p_{\text{meta}}(\mathcal{T})} \left[J(\theta^{*}(\mathcal{T},\bm{\Theta}))\right]\\
    &\text{over all} && \bm{\Theta} \in \mathbb{R}^{n}.
\end{aligned}
\label{eq:Metaobjective}
\end{equation}
Crucially, in the context of process control, meta-RL does \emph{not} aim to find a single controller that performs well across different plants. Note that $\theta^{*}$ in \cref{eq:Metaobjective} is the optimal weight vector in \cref{eq:RLobjective} as a function of a sampled MDP $\mathcal{T}$ and the meta-weights $\bm{\Theta}$. Meta-RL agents aim to simultaneously learn the underlying structure characterizing different plants and the corresponding optimal control strategy under its cost function. The practical benefit is that this enables RL agents to quickly adapt to novel environments.\par

There are two components to meta-learning algorithms: the model (e.g., the policy network) that solves a given task, and a set of meta-parameters that learn how to update the model \cite{bengio2013optimization, andrychowicz2016learning}. Due to the shared structure among tasks in process control applications, we are interested in \emph{context-based} meta-RL methods \cite{duan2016rl, wang2016learning, rakelly2019efficient}. These approaches learn a latent representation of each task, enabling the agent to simultaneously learn the context and the policy for a given task. This contrasts with other recent RL work on PID tuning, such as \cite{2022pidrl}, where a process model is identified on a case-by-case basis and an RL agent is trained for a specific process.\par

Our method is similar to \cite{duan2016rl}: the main idea is to view \cref{eq:Metaobjective} as a single RL problem. For each MDP $\mathcal{T} \sim p(\mathcal{T})$, the meta-RL agent has a maximum number of time steps, $T$, to interact with the environment, called an \emph{episode}. As each episode progresses, the RL agent has an internal hidden state vector $z_t$ which evolves with each time step through the MDP based on the RL states the agent observes: $z_t = f_{\bm{\Theta}}(z_{t-1},s_t)$. The RL agent conditions its actions on both $s_t$ and $z_t$. An illustration of this concept is shown in \cref{fig:RL2}. Therefore, the purpose of the meta-parameters $\bm{\Theta}$ is to quickly adapt a control policy for a MDP $\mathcal{T} \sim p(\mathcal{T})$ by solving for a suitable set of MDP-specific parameters encoded by $z_t$. This is why this approach is described as meta-RL; rather than training a reinforcement learning agent to control a process, we are training a meta-reinforcement learning agent to find a suitable set of parameters for a reinforcement learning agent which can control a process. The advantage of training a meta-RL agent is that the final model is capable of controlling every MDP across the task distribution $p(\mathcal{T})$ whereas a regular RL agent could only be optimized for a single task $\mathcal{T}$.\par
Clearly, the key component of the above framework is the hidden state. This is generated with a recurrent neural network (RNN), a special neural network structure for processing sequential data. An RNN can be thought of as a nonlinear state-space system that is optimized for some objective. The RNN structure we use in practice is a gated recurrent network \cite{cho2014learning}. 


\section{Meta-RL for process control}
\label{sec:algorithm}

We apply the meta-RL framework to the problem of tuning proportional-integral-derivative (PID) controllers. The formulation can be applied to any fixed-structure controller, but we focus on PID controllers due to their industrial prevalence.

\subsection{Tasks, states, actions, costs}

The systems of interest are second-order plus time delay (SOPTD):
\begin{equation}
    G(s) = \frac{K}{(\tau_1 s + 1)(\tau_2 s + 1)} e^{-\theta s},
    \label{eq:foptd}
\end{equation}
where $K$ is the process gain, $\tau_1$ and $\tau_2$ are the system poles ($\tau_1 \geq \tau_2)$, $\theta$ is the time delay, and $s$ is the Laplace variable (not to be confused with $s_t$, which represents the RL state at time step $t$).\par
A PID controller has the form:
\begin{equation}
    C(s) = K_c \left (1 + \frac{1}{\tau_I s} + \tau_D s \right),
    \label{eq:PI standard form}
\end{equation}
where $K_c$, $\tau_I$, $\tau_D$ are tuning parameters. We also apply a low-pass filter to the derivative error term in the controller. Prior work on RL for PID tuning suggests an update scheme of the form \cite{lawrence2022deep}:
\begin{align}
    [K_c, \tau_I, \tau_D] &\leftarrow [K_c, \tau_I, \tau_D] + \alpha \nabla J([K_c, \tau_I, \tau_D])\\
    &=  [K_c, \tau_I, \tau_D] + \Delta [K_c, \tau_I, \tau_D]
\end{align}
where the RL policy is directly parameterized as a PID controller. We follow this update scheme and take the actions to be changes to the PID parameters $\Delta [K_c, \tau_I, \tau_D]$.\par

The MDP state contains the current PID parameters as well as the proportional setpoint error and the integral setpoint error calculated from the beginning of an episode, $t_0$, to the current time step, $t$. 
\begin{equation}
    s_t = \left[K_c, \tau_I, \tau_D, e_t, \int_{t_{0}}^{t} e_{\tau} d\tau \right]
    \label{eq:state definition}
\end{equation}
The RL agent is trained to minimize its discounted future cost interacting with different tasks. The cost function used to train the meta-RL agent is the squared error from a target trajectory, shown in \cref{eq:cost}. The target trajectory is calculated by applying a first order filter to the set point signal, shown in \cref{eq:filter}. The time constant of this filter is set to the desired closed-loop time constant, $\tau_{cl}$. A target closed-loop time constant is chosen for robustness and smooth control action following the PID tuning rules presented in \cite{skogestad2003simple} and shown in \cref{eq:cltimeconstat}. An $L1$ regularization penalty $\beta > 0$ on the agent's actions is also added to the cost function to encourage sparsity in the meta-RL agent's output and help the tuning algorithm converge to a constant set of PID parameters (rather than acting as a non-linear feedback controller and constantly changing the controller parameters in response to the current low-level state of the system).
\begin{equation}
    c_t = (y_{desired,t} - y_t)^2 + \beta_1 |\Delta K_c| + \beta_2 |\Delta \tau_I| + \beta_3 |\Delta \tau_D|,\label{eq:cost}
\end{equation}
where
\begin{align}
    Y_{desired}(s) &= \frac{y_{sp}}{\tau_{cl} s + 1}e^{-\theta s}\label{eq:filter}\\
    \tau_{cl} &= 2\tau_1 + \tau_2,~\text{with }\tau_1 > \tau_2\label{eq:cltimeconstat}
\end{align}
Comparing the RL state definition to the RL cost definition, we see similar trajectories through different MDPs will receive very different costs depending on the underlying system dynamics in the particular tasks being controlled. In order for the meta-RL agent to perform well on a new task, it needs to perform implicit system identification to generate an internal representation of the system dynamics.

The advantages of this meta-RL scheme for PID tuning are summarized as follows:
\begin{itemize}
    \item Tuning is performed in closed-loop and without explicit system identification.
    \item Tuning is performed automatically even as the underlying system changes.
    \item The agent can be deployed on novel ``in distribution'' systems (e.g. systems within the task distribution $p(\mathcal{T})$) without any online training. Additionally, nearly any system's data can be modified to be ``in-distribution'' \cite{mcclement2022meta2}.
    \item The meta-RL agent is a single model that is trained once, offline, meaning one does not need to specify hyperparameters on a task-by-task basis.
    \item The meta-RL agent's cost function is conditioned on the process dynamics and will produce consistent closed-loop control behaviour on different systems.
\end{itemize}
This approach is not limited to PID tuning. It can also be applied to other scenarios where the model \emph{structure} is known. The agent then learns to behave near-optimally inside each task in the training distribution, bypassing the need to identify model parameters and only train on that instance of the dynamics.

\subsection{RL agent structure}

\begin{figure}
    \center
    \includegraphics[scale=0.45]{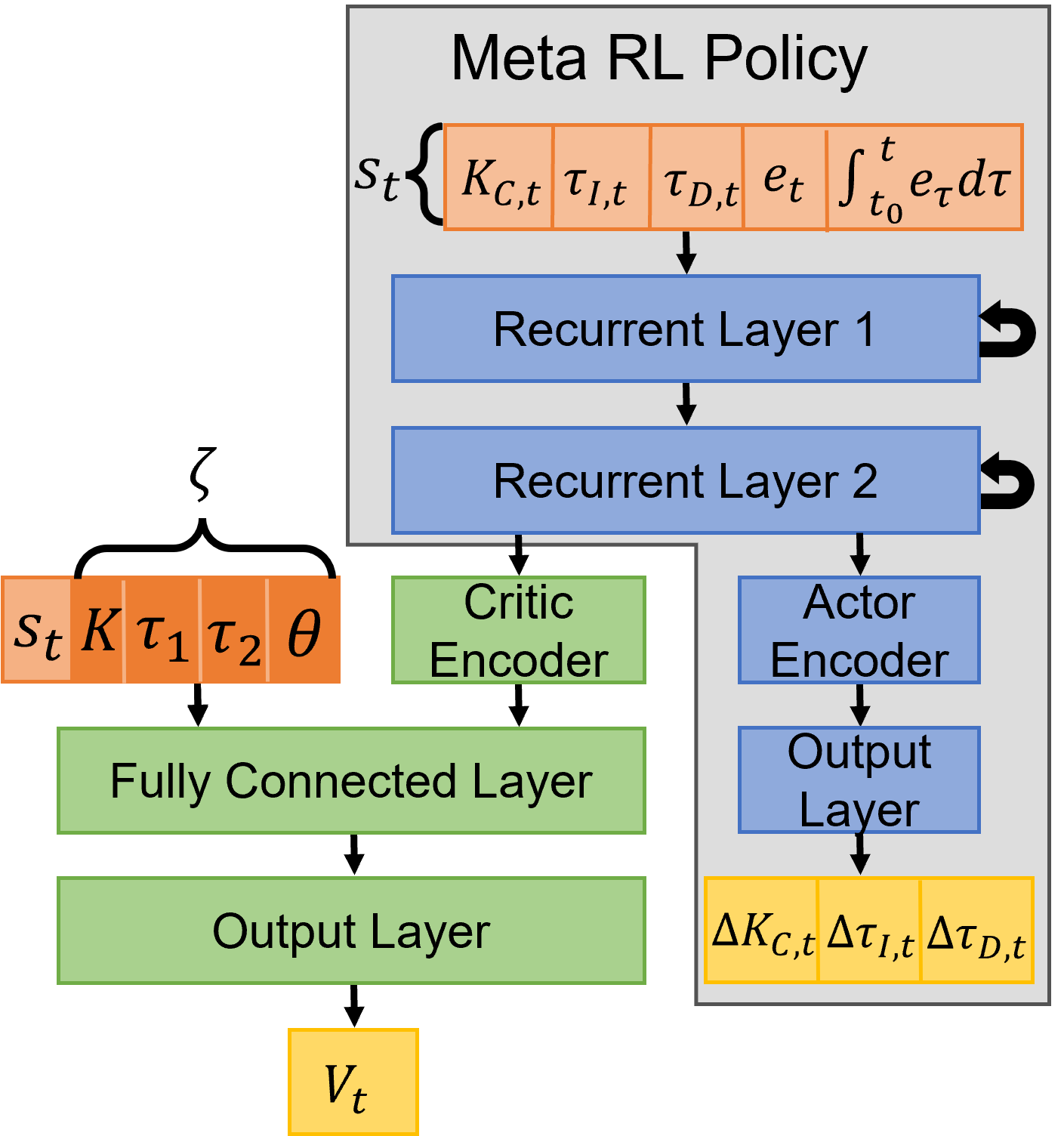}
    \caption{The structure of the RL agent. The control policy used online is shown in the grey box while the critic used during offline training is shown in green.}
    \label{fig:controller-structure}
\end{figure}

The structure of the meta-RL agent is shown in \cref{fig:controller-structure}. The grey box shows the ``actor'', i.e., the part of the agent used online for controller tuning. Through interacting with a system and observing the RL states at each time step, the agent's recurrent layers create an embedding (hidden state) which encodes information needed to tune the PID parameters, including information about the system dynamics and the uncertainty associated with this information. These embeddings essentially represent process-specific RL parameters which are updated as the meta-RL agent's knowledge of the process dynamics changes. Two fully connected layers use these embeddings to recommend adjustments to the controller's PID parameters. The inclusion of recurrent layers is essential for the meta-RL agent's performance. Having a hidden state carried between time steps equips the agent with memory and enables the agent to learn a representation of the process dynamics. A traditional feedforward RL network would be unable to differentiate between different tasks and would perform significantly worse. This concept is demonstrated in \cite{mcclement2021meta}.\par

Outside of the grey box are additional parts of the meta-RL agent which are only used during offline training. The ``critic'' (shown in green) is trained to calculate the value (an estimate of the agent's discounted future cost in the current MDP given the current RL state). This value function is used to train the meta-RL actor through gradient descent via the PPO algorithm.\par

A unique strategy we use to improve the training efficiency of the meta-RL agent is to give the critic network access to ``privileged information'', defined as any additional information outside the RL state and denoted as $\zeta$. In addition to the RL state, the critic conditions its estimates of the value function on the true process parameters ($K$, $\tau_1$, $\tau_2$, and $\theta$), as well as the deep hidden state --- the hidden state of the second recurrent layer --- of the actor. Knowledge of a task's process dynamics, as well as knowledge of the actor's internal representation of the process dynamics through its hidden state, allows the critic to more accurately estimate the value function, which improves the quality of the surrogate objective function used to train the actor and the training efficiency of the algorithm \cite{mcclement2022meta2}. Equipping the critic with this information also allows it to operate as a simpler feedforward neural network rather than a recurrent network like the actor.\par

The privileged information given to the critic network may at first appear to conflict with the advantages of the proposed meta-RL tuning method, since the critic requires the true system parameters and much simpler tuning methods for PID controllers exist if such information is known. However, this information is only required during \emph{offline} training. The meta-RL agent is trained on simulated systems with known process dynamics, but the end result of this training procedure is a meta-RL agent that can be used to tune PID parameters for a real process \emph{online} with no task-specific training or knowledge of the process dynamics. The portion of the meta-RL agent operating online contained in the grey box only requires RL state information --- process data --- at each time step.

\subsection{Training procedure}

The meta-RL agent is trained by uniformly sampling $K$, $\tau_1$, $\tau_2$, and $\theta$ to create a SOPTD system and initializing a PID controller with $K_c=0.05$, $\tau_I=1.0$, $\tau_D=0.2$. The state of the system is randomly initialized near zero by sampling from $\mathcal{N}(0,0.1)$ and the set point is switched between $1$ and $-1$ every $11$ units of time. The meta-RL agent has no inherent time scale and so we keep the units of time general to highlight the applicability of the proposed PID tuning algorithm to both fast and slow processes (allowing time constants on the order of milliseconds or hours).\par

We consider the following sets of SOPTD model parameters in our experiments: $K$ and $\tau_1$ are sampled from the interval $[0.25, 1.0]$; The ratios $\frac{\tau_2}{\tau_1}$ and $\frac{\theta}{\tau_1}$ are sampled from the interval $[0.0, 1.0]$. In \cite{mcclement2022meta2} we demonstrate how training across this range of parameters can be quite versatile in practice using data augmentation to adjust the dynamics of most second order processes into this distribution from the meta-RL agent's perspective.
The PPO algorithm is adapted from Open AI's ``Spinning Up'' implementation and modified to accommodate the inclusion of a recurrent neural network and distribution of control tasks \cite{SpinningUp2018}. 


\section{Results}
\label{sec:results}

\begin{figure}
    \centering
    \leftskip-0.08in
    \includegraphics[scale=0.4]{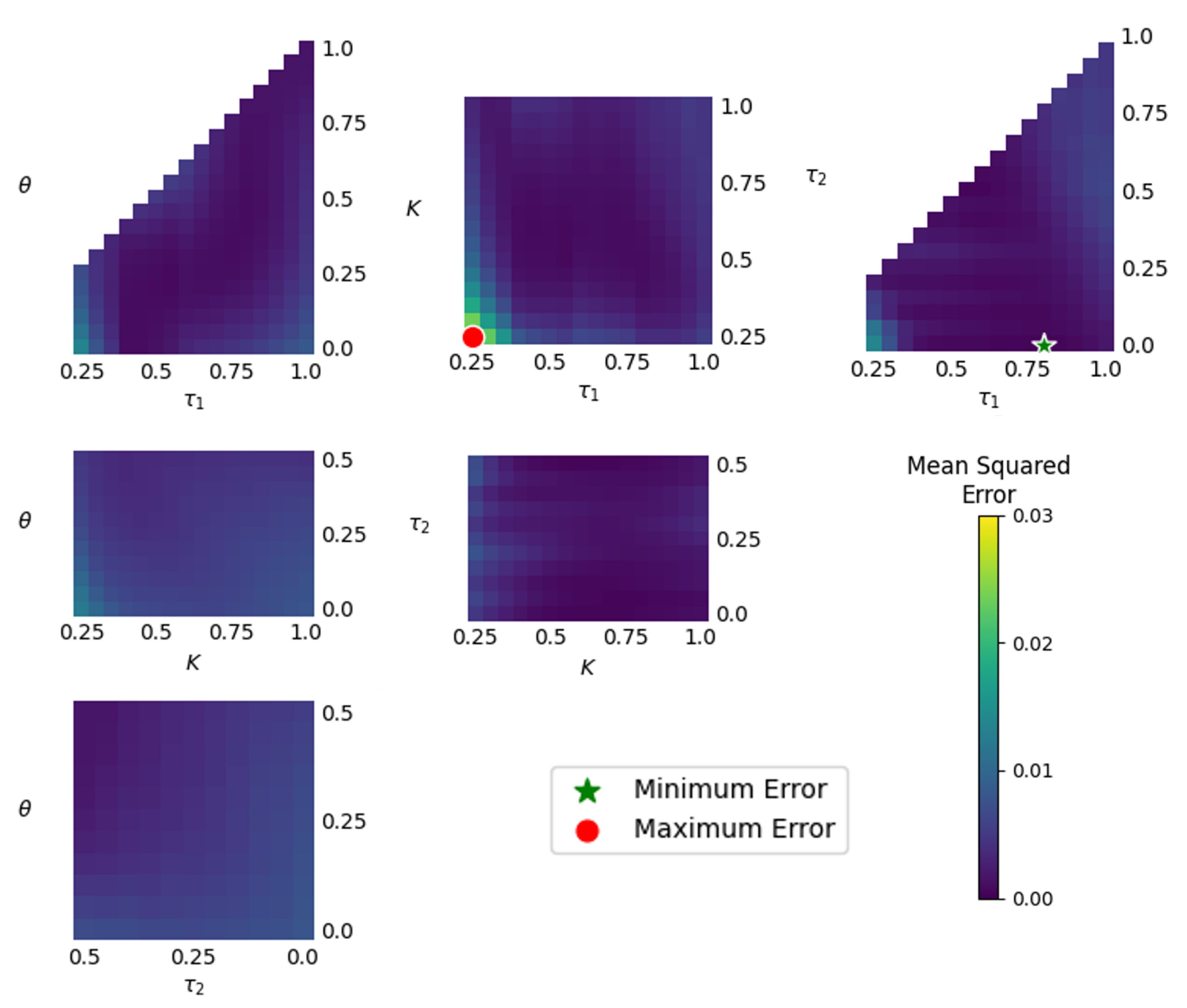}
    \caption{Performance of the Meta-RL agent across $p(\mathcal{T})$. By default, $K=0.5$, $\tau_1=0.5$, $\dfrac{\tau_2}{\tau_1}=0.5$, $\dfrac{\theta}{\tau_1}=0.5$. The color represents the mean squared error from a target trajectory during a step response, as shown in Fig. 4.}
    \label{fig:performance}
\end{figure}

After training, the meta-RL model is tested on a range of processes within $p(\mathcal{T})$. To visualize the performance of the meta-RL agent, we plot the mean squared error from the target trajectory when the set point changes 2 units for different SOPTD systems after the agent has had time to adapt. Because the set of possible processes is four-dimensional, we hold two parameters constant at their mean values to plot how the performance varies as the other two parameters are manipulated. The performance is shown in \cref{fig:performance}.

In \cref{fig:performance} we see the agent replicates the desired trajectory quite accurately and consistently across $p(\mathcal{T})$. Performance is significantly worse when the process gain and dominant pole both have small magnitudes. These systems require the largest changes to the initial PID parameters and the regularization added in \cref{eq:cost} to changes in PID parameters results in a slightly under-tuned controller in these cases. 

\begin{figure}
    \centering
    \includegraphics[scale=0.5]{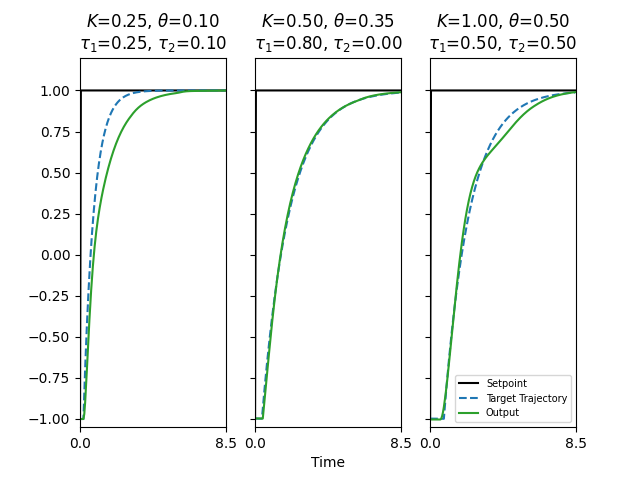}
    \caption{A demonstration of the trajectory-tracking of PID tunings produced by the meta-RL agent ability to track the target trajectory in the worst case (left), best case (center) and with significant second order dynamics (right).}
    \label{fig:trajectory}
\end{figure}

To further illustrate the meta-RL agent's PID tuning performance, a set point change on the best and worst-case systems from \cref{fig:performance} is shown in \cref{fig:trajectory}. A third trajectory is shown to illustrate the performance on systems which have more prominent second-order behaviour ($\tau_2 \approx \tau_1)$.

\begin{figure}
    \centering
    \leftskip-0.28in
    \includegraphics[scale=0.5]{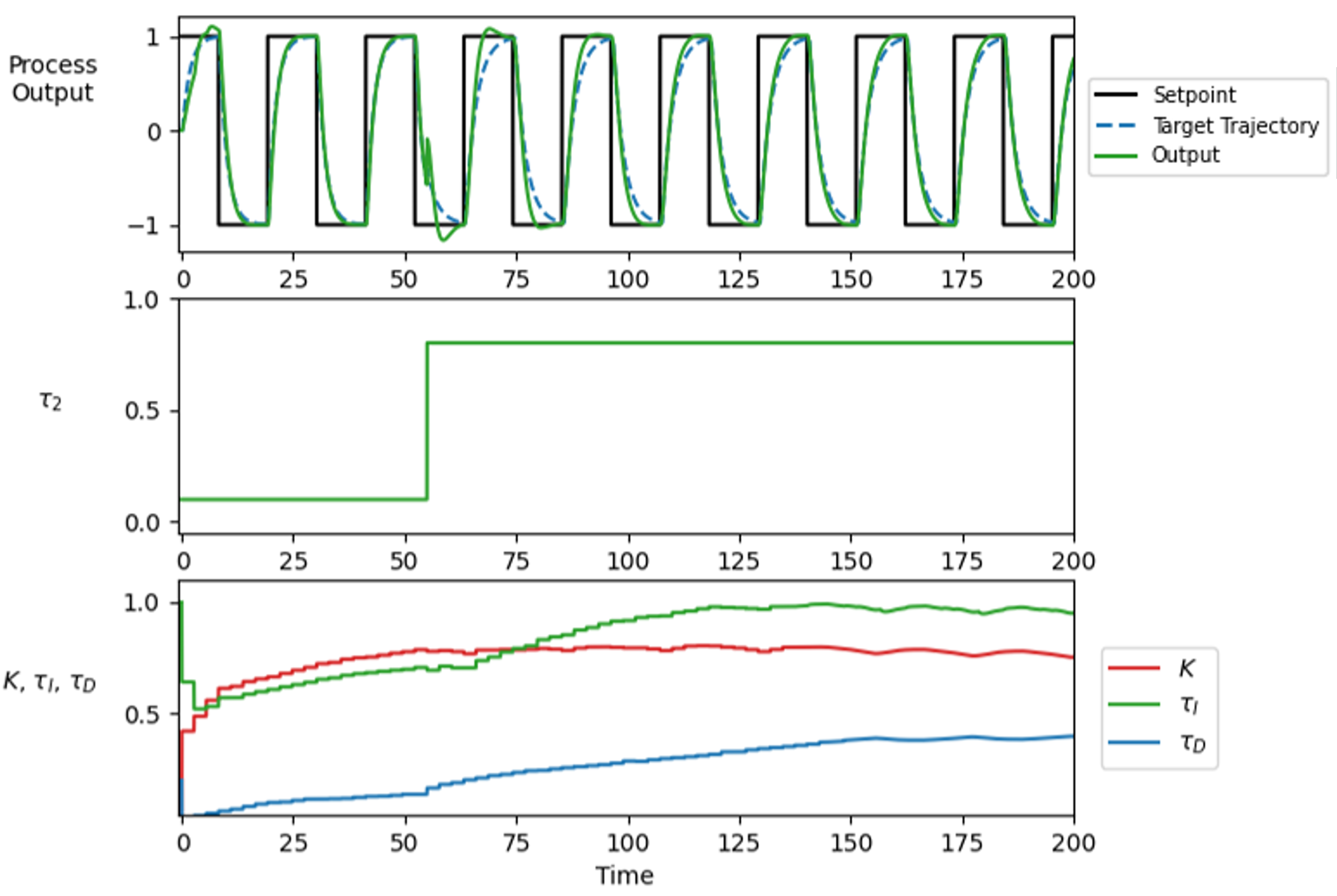}
    \caption{A demonstration of the initial PID tunings produced by the meta-RL agent on the system $k=0.5,~\tau_1=0.8,~\tau_2=0.1,~\theta=0.05$ as well as the meta-RL agent's ability to adapt the PID parameters when the process dynamics change ($\tau_2$ is increased to 0.8 to simulate more prominent second-order behaviour).}
    \label{fig:adaptation}
\end{figure}

\cref{fig:adaptation} demonstrates the meta-RL agent's initial PID adaptation when initialized on a new system. To illustrate the meta-RL agent's ability to adapt its tunings when the system dynamics change, more prominent second-order behaviour is introduced by increasing $\tau_2$ and we see the meta-RL agent is able to recognize and adapt the PID parameters in response to this change in the process dynamics.

\begin{figure}
    \centering
    \includegraphics[scale=0.5]{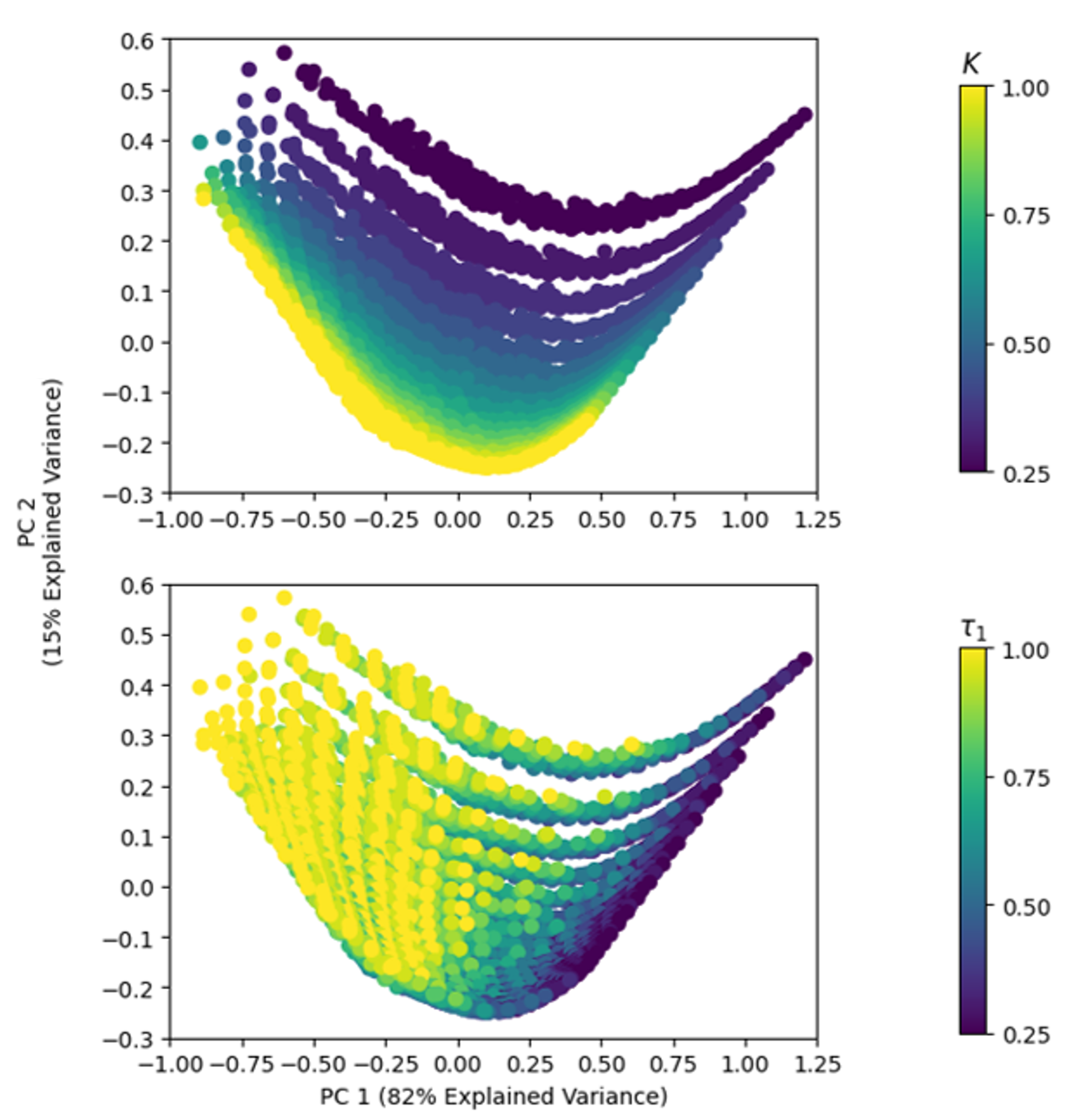}
    \caption{Principal component visualization of the meta-RL agent's deep hidden state after adapting to various processes.}
    \label{fig:pca1}
\end{figure}

Next, we analyze the hidden state of the meta-RL agent after adapting to a process. We reduce the hidden state to two principal components (PCs) for two-dimensional visualization in \cref{fig:pca1}. We see the hidden state of the meta-RL agent does encode information about the process gain and dominant pole $\tau_1$, confirming our hypothesis that the agent is performing implicit system identification.

\begin{figure}
    \centering
    \includegraphics[scale=0.57]{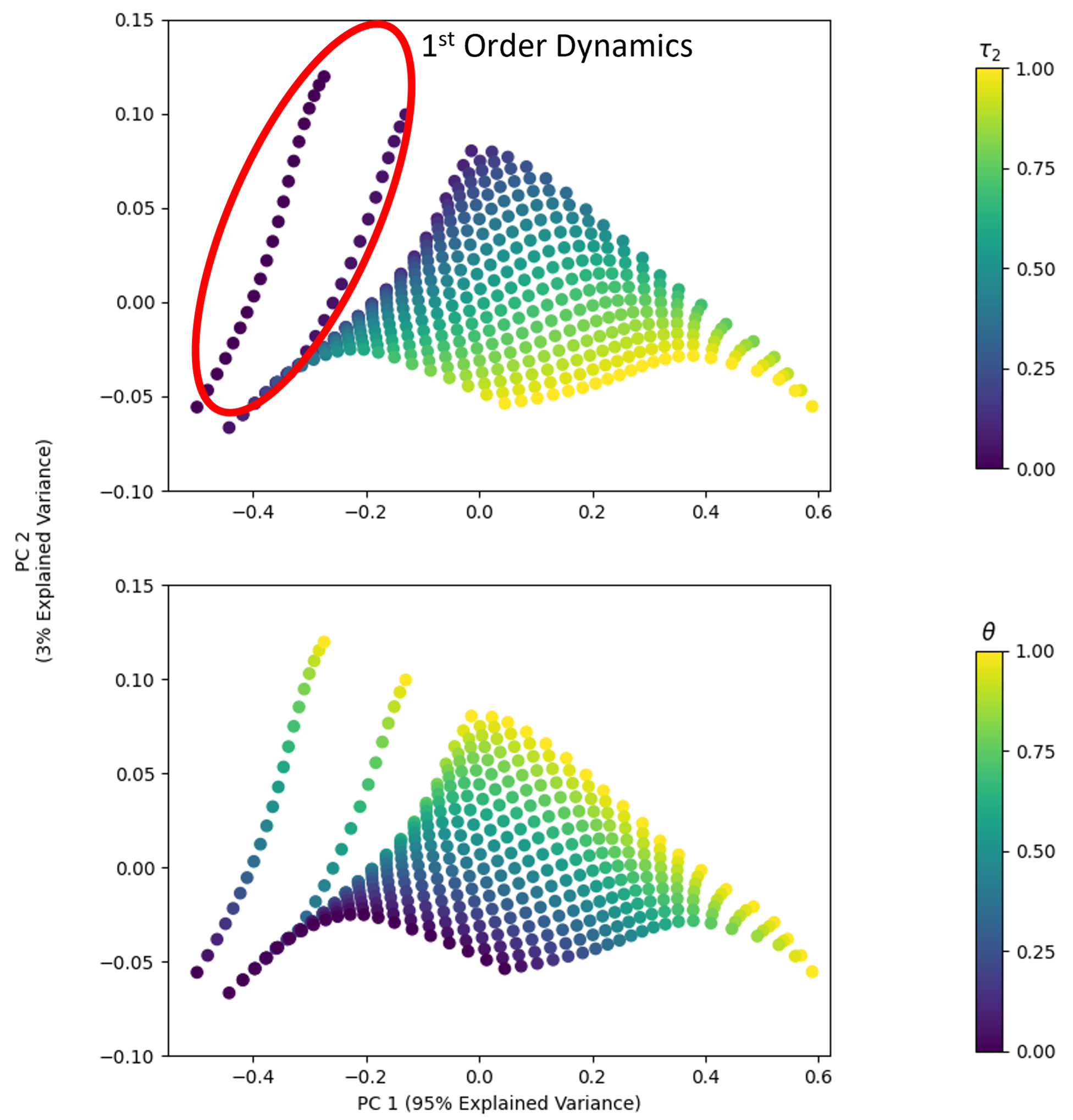}
    \caption{Principal component visualization of the meta-RL agent's deep hidden state after adapting to various processes with $K=1$ and $\tau_1=1$.}
    \label{fig:pca2}
\end{figure}

In \cref{fig:pca2}, we recalculate the principal components when $K$ and $\tau_1$ are held constant to better visualize whether the hidden state contains information about $\tau_2$ and $\theta$ as well. From the pattern, we see information about $\tau_2$ and $\theta$ are encoded in the hidden states. Another interesting result is the differentiation between processes with first or near first-order dynamics ($\tau_2 = 0$ or $\tau_2 \ll \theta$) from other processes within $p(\mathcal{T})$.

\section{Conclusion}
\label{sec:conclusion}

In this work, we presented a meta-RL model capable of tuning fixed-structure controllers in closed-loop without explicit system identification and demonstrated our approach using PID controllers. The tuning algorithm can be used to help automate the initial tuning of controllers or maintenance of controllers by adaptively updating the controller parameters as process dynamics change over time. Assuming the magnitude of the process gain and time constant are known, the meta-RL tuning algorithm can be applied to any system which can be reasonably approximated as a SOPTD system. The present work focuses on lag dominant SOPTD systems ($\tau_1>\theta$), however the results could be extended to dead-time dominant systems by expanding the task distribution $p(\mathcal{T})$ used during training. \par

A major challenge of applying RL to industrial process control is sample efficiency. The meta-RL model presented in this work addresses this problem by training a model to control a large distribution of possible systems offline in advance. The meta-RL model is then able to tune fixed-structure process controllers online with no process-specific training and no process model. There are two key design considerations which enable this performance. First is the inclusion of a hidden state in the RL agent, giving the meta-RL agent a memory it uses to learn internal representations of the process dynamics through process data. Second is constructing a value function which uses extra information in addition to the RL state.\par


\bibliographystyle{IEEEtran}
\bibliography{main}

\end{document}